# A Mathematical Framework for Superintelligent Machines

**Daniel J. Buehrer (Lifetime Senior Member, IEEE)**
Address: Institute of Computer Science and Information Engr., National Chung Cheng University, Chiayi 621, Taiwan

(e-mail:  dan@cs.ccu.edu.tw)

**ABSTRACT** We describe a class calculus that is expressive enough to describe and improve its own learning process.  It can design and debug programs that satisfy given input/output constraints, based on its ontology of previously learned programs.  It can improve its own model of the world by checking the actual results of the actions of its robotic activators.  For instance, it could check the black box of a car crash to determine if it was probably caused by electric failure, a stuck electronic gate, dark ice, or some other condition that it must add to its ontology in order to meet its sub-goal of preventing such crashes in the future.  Class algebra basically defines the *eval*/*eval*$^{-1}$ Galois connection between the residuated Boolean algebras of
   1.   equivalence classes and super/sub classes of class algebra type expressions, and
   2.   a residual Boolean algebra of biclique relationships.

It distinguishes which formulas are equivalent, $\rightarrow$, $\leftarrow$, or unrelated, based on a simplification algorithm that may be thought of as producing a unique pair of Karnaugh maps that describe the rough sets of maximal bicliques of relations.  Such maps divide the *n*-dimensional space of up to $2^{n-1}$ conjunctions of up to *n* propositions into clopen (i.e. a closed set of regions and their boundaries) causal sets.  This class algebra is generalized to type-2 fuzzy class algebra by using relative frequencies as probabilities.  It is also generalized to a class calculus involving assignments that change the states of programs.

### I.   INTRODUCTION

This paper describes what Pedro Domingos [1] has called "The Master Algorithm".  This algorithm will probably be stored on a cloud along with personalized ontologies of classes, binary relations, and programs (i.e. services).  It will control ontologies of chat bots and robots of individuals, communities, and institutions.  These bots will provide sensor/activator data for the personalized superintelligent machine (sim) ontologies to learn how to effectively achieve their goals.

The many tribes of Artificial Intelligence learning [2] can be unified by this algorithm, which we call *class calculus*. It is a dynamic version (i.e. with assignments) of *class algebra*, a theory that satisfies the axioms of *causal sets* [3], *relation algebra* [4-8], and *residuated Boolean algebras* [9,10].  The *constructionists' deep learning neural* and *capsule networks*, the *symbolists' inverse deduction*, the *evolutionists' genetic programming*, *Bayesian inference*, and *analogical reasoning*'s *support vectors* can all be described by sub-algebras within the *conceptual framework* of this *class algebra*.

Deep Learning neural networks have proven to be very successful at learning in narrow domains such as playing games or driving a car. However, there has been some difficulty in combining these *black box AI* programs into a more general framework. *Class calculus* offers a way out of this dilemma.  Rather than learning only from data, class calculus learns from <u>hierarchical descriptions</u> that are created from the data. That is, *mathematical patterns* at *various levels* of the data are recognized and used to *compress* and *generalize* the data, similar to the way that *Hinton's capsule networks* use layered descriptions of *edge convolutions*, region colors and textures, and *poses* and distances among various parts to recognize an object such as a face.

 *Class algebra (CA)* is a *decidable extension* of *causal set theory*, *Tarski's relation algebra* [4-8], and *residuated Boolean algebras* of both sets and their logical descriptions. Its logic inference algorithm creates an *IS-A hierarchy* for classes, binary relations, and programs. This hierarchy is based on expressing every relation as a union of *maximal bicliques*.  Each of a relation's max bicliques becomes a domain and range class of a subrelation. So, CA may be thought of as the Boolean algebra of *equivalence classes* of binary relations. The elements of these equivalence classes all have the same *type*. These *homogeneous collections*, (e.g. sorted sets, graphs, and hash tables) are described by sub-algebras of class algebra.

The replacement operations of the class algebra simplification process always halt with *a unique simplest form*. This form can be thought of as a pair of *Karnaugh maps* for the intents of sets S and ~ -S.

However, to get *superintelligence*, the computational system must be able to look at its own computations and do *induction*, and also find mathematical *morphisms* and other simplifications in its proofs. For example, in the *pigeon-hole problem*, the system will have to learn that the number of "unplaced" pigeons and "empty" pigeon holes can be used to simplify the descriptions of the sets of all possible placements of pigeons into holes, since the solutions are *isomorphic* under *permutations* of the lists of pigeons and holes. Such matching and counting arguments are just the beginning of being able to use intelligent mathematical tools to simplify proofs. But finiteness of the number system (e.g. *IEEE 32-bit integers* or *ASCII* integers) must be assumed to prevent *Gödel's incompleteness theorem* from applying to the computations. For finite descriptions, class algebra is complete for *direct proofs*, but *inductive proofs* require the guessing of strong enough *loop invariants* based on previously seen "similar" instances of such problems. *Similarities* and *analogies* can also lead to logical hypotheses that are testable by using *the scientific method* of checking that there exist instances that satisfy any new class definition's intent.

We give a formal definition of class algebra in Section 2 and briefly sketch a *resolution-based reasoning algorithm* for it in Section 3. We also define the semantics of class logic in terms of matrices in Section 4. We describe the type-2 fuzzy version of class algebra and its reasoning algorithm in Section 5. Section 6 extends class algebra to a class calculus for modeling changes of state among independent *services* that send *messages* to each other, and we compare this learning model to other rule-based computational models. Section 7 proposes some standardized names for the level 1 IS-A hierarchy and its attributes. Finally, in Section 8 we discuss some philosophical and *moral issues of super-intelligent machines (sims)*. We leave to future papers the use of this algorithm for solving the problems in SAT *satisfiability problem sets* (e.g. the pigeon-hole problem) and the verification that this meta-theory applies to various new theories of learning. As for any theory, the class calculus framework can perhaps not be extended to learn some undecidable theories (e.g. *non-algebraic real numbers* and *first-order logic*), but that is possibly not important since the physical universe seems to be a finite system of about $10^{150}$ quanta.

## II. Class Algebra Definition

Class algebra is an algebra <eval/2, ⊆/2, ∪/2, ∩/2, ·/2, ~/1, -/1, $^{-1}$/1, *1*/1, *0*/1, @/1, *U*/0, ∅/0, *N*/0, Σ/0>, where the infix operators are right associative and are listed in increasing order of precedence.

Class algebra involves names *N* that are in *1-1 onto* correspondence with nodes of a *partial order*. According to a complete order of CA expressions, one CA expression is chosen as "the" simplified *name* of each node, although the other names may be saved in *aliases*. The partial order is called the *IS-A hierarchy*. As for any partial order, it has an implicit lattice. The ∅ and *U* are the bottom and top nodes of the IS-A partial order and its lattice, and the join ∪ and meet ∩ operators represent both the union/intersection operators of Boolean sets and the logical "or" and "and" operators of Boolean logic. They also represent the join/meet operators of CA rough sets and rough relations.

Each node in CA's IS-A hierarchy has both a logical *intent* and a causal set *extent*. The extent is the set of top nodes below both this class and the "SingularObject" class in the IS-A hierarchy. Each intersection class $\cap_j Y_j$ intersects the extents of the $Y_j$ and conjoins their *intents*.

Properties of classes, relations, and programs are stored in their *intents*. The intent is an implicit conjunction of the intents of all nodes above it in the IS-A hierarchy. Intents use a *Tseytin conjunctive normal form (CNF)* [11], where each disjunction of a Boolean expression is given a unique "hidden" class name $h_i$. The intent of a union class $h_i \leftrightarrow \cup_j Y_j$ is a "simplified" disjunction of the class intents of the $Y_j$ in the IS-A hierarchy. The simplification process can be described as finding the closure of the CA intent under resolution [12] and subsumption. Each union class "extent" is the union of all $Y_j$ extents. Resolution involves xy∨~xz→yz, where the subsets of x and ~x have an empty intersection. Subsumption uses the absorption law xy∨x=x.

The intent constrains the range of max bicliques where this class is the domain of these bicliques. The intent (i.e. logical description) of a class definition involves propositions of the form "*expr*⊆*C*" where *expr* is the name of the *left adjoint functor* of a binary relation or its *inverse*, *expr*$^{-1}$, or a CA expression involving relation compositions of these adjoint functors. *C* is the name of a subclass where [*C*,~-*C*] gives a lower/upper bound of *expr*'s range.

A 4-valued, *strongly typed* Boolean algebra is formed by adding a Boolean complement operator (~) and by adding a *universe U* to the top of the class hierarchy and *empty class* ∅ to the bottom, with an intent that conjoins all *literals* (i.e. p, -p, ~p, ~-p for all *propositions* p of the form "*expr*⊆*C*". That is, class algebra also uses a pseudo-complement (-) relation, with ~-r forming an *upper bound* on the possible edges of relationships or the permissible objects in a class extent. This theory of rough sets is then extended to *type-2 fuzzy sets* by counting the relative number of objects in the *rough sets*.

Although *residuated Boolean algebras* use the dot

operator of *monoids*, class algebra restricts this to the dot operator from *group* theory. It must satisfy *associativity*, existence of *left/right identities*, and the group *inverse* $(xy)^{-1}=y^{-1}x^{-1}$. It can thus represent *concatenation* of strings in *formal language theory* (where letters are their own reverses), *inner* and *outer* products of *matrices*, *functor composition* in *category theory*, "hasPart" relations of *aggregates*, nesting of regions in space-time, the *cons* constructor of lists, and successive transitions in the *state spaces* of *automata*. That allows the exploration of finite state spaces, as for games like GO or for *context sensitive* replacement operations.

The dot operator of Tarski's relation algebra [4-8] is that of a monoid, but, as mentioned, CA also assumes that the monoid is a group that satisfies the law $(R·S)^{-1}=S^{-1}·R^{-1}$ for all relations R and S. R·S represents *compositions* of similarly typed collections, such as binary relations (i.e. over $U$), strings, lists, arrays, or hash sets. The node names $N$ include CA expressions of the form "@" Classname (".") adjointFunctorName)*. For class calculus, such node names can also be followed by the *outfix* operators [SubscriptList] for vectors, (ArgList) for services, or {relativeCAexpr} to select a subclass of a class expression. The relativeCAexpr is simply a class algebra expression where unambiguous prefixes of names can be left off. These environments can be statically nested, just like "with" statements of a language like Pascal.

The names $N$ can include valid *Extensible Resource Identifiers*, (XRIs). Nested ":" infix operators can be used for a *dynamically nested Lisp-like environment* of names and their values. For example, urn:isbn:0-486-7557-4 is the XRI (and also *URN*) a specific edition of *Romeo and Juliet.*

For the sake of simplicity, we assume that the alphabet Σ is the *UTF-8 character set*, since this includes names for classes and relations in languages such as Chinese and Arabic. UTF-8 includes the *ASCII characters*. The IEEE encoding for 32-bit integers and *floating point numbers* could also be encoded with appropriate rules for addition, subtraction, multiplication, and division for arbitrarily long strings of digits and a decimal point, obtaining an implementation of Tarski's theory of decidable real closed fields [6].

Class algebra (CA) is a *relation algebra* [4-8] <$L$, ∪, ∩, ~, ∅, $I$, ·, $^{-1}$> for a *bounded lattice* $L$=<$N$/0, ∪/2, ∩/2, $U$/0, ∅/0> with CA expressions $N$ labeling the nodes. CA includes the laws of the *residuated Boolean algebra*, <∪/2, ∩/2, ~/1, $2^{N×N}$/0, $N$×$N$/0, ⊆/2> for extents of relations, the *Heyting algebras* <∪/2, ∩/2, $-_C$/1, $2^{C×C}$/0, $C$×$C$/0> for sub-algebra extents with $C$ in $N$, the *group* <⊆/2, ·/2, $^{-1}$/1, @/1, $N$/0> for strings, lists, and sorted sets, and the *residuated lattice* <$L$, → /2, ·/2, $I$/0> which describes the search space for the *context-sensitive reductions* that are used to describe logical deduction steps.

The pseudo-complement $(-)_C$ is rather unique to class algebra, and it is used to represent a complement operator for the sub-algebra with a top "base" node $C$ somewhere in class algebra's IS-A hierarchy. The "true" Boolean complement "~" is taken with respect to all nodes in the IS-A hierarchy (i.e. $N$), but the pseudo-complement is relative to a given "base" superclass of the given node. For example, "$-_{Animal}$Dog" is any Animal that is not a dog, whereas ~Dog includes stones and abstract ideas.

The class Dog can correspond to an ambiguous, open-world dictionary definition of "dog". If a person runs across the phrase "hot dog", that phrase could be added to this English dictionary for the word "dog". This *sense* of the word, however, will have its own class, which is not a subclass of Animal. Both Dog's intent and extent will indicate that Dog is in the algebra of the union of Food and Animal. The elements of $~-_{Animal∪Food}$Dog are in the intersection of Food, Animal, and Dog. IBM's Watson is very good at finding such ambiguous English word senses.

Type-2 fuzzy class algebra uses the ranges [size(X)/$n$, ($n$-size(-X))/$n$] where $n$=size($N$) for any class algebra expression X. Although size(X)/$n$ is a lower bound on the relative frequency, the upper bound may be less than the lower bound if X∪-X⊃X.base (i.e. if X∩-X ≠∅).

If the pseudo-complement operator satisfies the laws of a sub-Boolean algebra, where x∧-x=∅$_x$, x∨-x=x.base, and where -~x=~-x, then the IS-A hierarchy contains a nested sub-algebra that describes a strongly-typed collection. Here, x.base (i.e. C) is the "relative universe" for the pseudo-complement operator $-_C$. These sub-algebras can have their own names for their bottom class, such as the empty string ε, the empty list [], or the empty set {} of some type. For example, the simplification process of class algebra expressions could use the algebra of either sorted lists or hash sets of propositions to simplify intents.

The concatenation operator (·/2) of a group (i.e. with R·R$^{-1}$=R.domain, R$^{-1}$·R=R.range and R·$I$=R=$I$·R) is associative but not necessarily commutative. The inverse $(R·S)^{-1}$ = $S^{-1}·R^{-1}$ simply reverses a list, string, or composition of binary relations. The dot operator can represent the *cons* list constructor of LISP and most computer languages. It can also be used to represent a string of operators, which are binary relations between states. For example, strings of operators produce the search spaces of *finite state automata* and their *hierarchical counterparts in UML*, *Earley state transitions*, *context-sensitive reductions* (over Σ$^+$/0 phrases (i.e. sense/activator signals) or hidden class names of unions of phrases (i.e. *nonterminals*)).

We use the notation of the logical operator → instead of the more common ≤ and ≥ symbols or ▷ and ◁ symbols of residual lattice theory. The concatenation operator can replace ∩ if the CA expression involves operators with side effects, in which case concatenation acts like an "and-then" operator. Nested → have the effect of "or-else" operators.

A shorthand notation can allow logic replacement operators to produce a new state by adding (+) and deleting (-) some propositions when going from the previous state to the new state. Like an intent, a state is simply a sorted conjunction of propositions of the form (adjointFuntor⊆C) for a class name C. By default, the name of the left adjoint functor is the same as the relation name R, while the right adjoint functor's name is the inverse relation's name (i.e. $R^{-1}$ has the default name "inv_R"). All adjoint functors have inverses, so search graphs can be run backwards to find the set of states which could lead to a given output state. Notice that relations are a union of max bicliques, and all max biclique domains and ranges may be given explicit names in the IS-A hierarchy (e.g. @Color{ @Light ∩ (@Red∪@Green)}). Such class algebra "selector" expressions correspond to the *Roles* of *description logics*.

Finally, the interpret relation *eval* is usually invoked from its adjoint functors $^{eval}$ or its inverse $^{describe}$. It is the relation that converts between the intent and extent of each class. The $^{describe}$ adjoint functor looks for previously seen class intents with the given extent, but it can also describe previously unseen extents via "minimal" Boolean unions of existing classes. Any extent has a finite number of such minimal expressions. For example, pq∪pr∪qr, with 5 operators, can also be represented by either p(q∪r)∪qr or pq∪(p∪q)r, each with 4 operators. Class algebra uses a decision procedure to prove that these formulas are in the same equivalence class.

**Table 1**
**Properties of Relationships**

| *R* is | If and only if: |
|---|---|
| Functional | $R˘·R ≤ I$ |
| Left-total | $I ≤ R·R˘$ ($R˘$ is surjective) |
| Function | functional and left-total. |
| Injective | $R·R˘ ≤ I$ ($R˘$ is functional) |
| Surjective | $I ≤ R˘·R$ ($R˘$ is left-total) |
| Bijection | $R˘·R = R·R˘ = I$ (Injective surjective fcn) |
| Transitive | $R·R ≤ R$ |
| Reflexive | $I ≤ R$ |
| Coreflexive | $R ≤ I$ |
| Irreflexive | $R ∧ I = 0$ |
| Symmetric | $R˘ = R$ |
| Antisymmetric | $R ∧ R˘ ≤ I$ |
| Asymmetric | $R ∧ R˘ = 0$ |
| Total | $R ∨ R˘ = 1$ |
| Connex | $I ∨ R ∨ R˘ = 1$ |
| Idempotent | $R·R = R$ |
| Preorder | *R* is transitive and reflexive. |
| Equivalence | *R* is a symmetric preorder. |
| Partial order | *R* is an antisymmetric preorder. |
| Total order | *R* is a total partial order. |
| Strict partial order | *R* is transitive and irreflexive. |
| Strict total order | *R* is a connex strict partial order. |
| Dense | $R ∧ Iˉ ≤ (R ∧ Iˉ)·(R ∧ Iˉ)$. |

Since class algebra is based on Tarski's relation algebra, what kind of properties about relations can be determined by looking at its relationships? Table 1 gives an answer. This table is taken from Wikipedia, since these definitions have not been collected into a book or journal paper yet.

In Table 1, the notation R˘ is used for $R^{-1}$ (i.e. the inverse, converse, transpose relation), and *I-* is used for the complement of the identity relation, where the identity relation can be thought of as the incidence matrix with 1's on the diagonal, and 0's elsewhere, for rows and columns containing all subsets of the *n* names *N*. That is, before simplification, there can be up to $2^n$ rows and columns in the identity relation *I*, although the complete relations *1* and *0* simplify to a 1✕1 matrix.

The residuals of Boolean algebra correspond to the elements below a given class (i.e. the classes whose intents all would → the intent of the given class). That is, a residuated Boolean algebra can be defined as follows:

A **residuated Boolean algebra** is an algebraic structure (*L*, ∧, ∨, ~, 0, 1, •, *I*, \, /) such that
(i) (*L*, ∧, ∨, •, *I*, \, /) is a residuated lattice, and
(ii) (*L*, ∧, ∨, ~, 0, 1) is a Boolean algebra.

Here, the residuals can be defined as follows:
  x(R\S)y ≡ ∀z∈X  zRx→zSy
  y(S/R)x ≡ ∀z∈X  xRz→ySz
For extents, the set *X* of subsets of *N*✕*N*, of size $2^{N^2}$, is made a Boolean algebra as usual with ∩, ∪ and ~ relative to the $2^{N^2}$ edges, and is also made a monoid with relation composition. The monoid unit *I* is the identity relation $\{(x,x)| x ∈ 2^{N^2}\}$. For intents, the right residual R\S is defined by x(R\S)y if and only if for all *z* in *X*, zRx implies zSy. Dually the left residual S/R is defined by y(S/R)x if and only if for all *z* in *X*, xRz implies ySz. Notice that the dot (composition) operators do not increase the size of the space $2^{N^2}$ since relation compositions R·S are unions of bicliques from equivalence classes in the domain of *R* and the range of *S*.

An equivalent signature of lattice *L* better suited to the relation algebra application is (*L*, ∧, ∨, ~, 0, 1, •, *I*, ▷, ◁) where the unary operations x\ and x▷ are inter-translatable in the manner of De Morgan's laws via
x\y = ~(x▷~y),   x▷y = ~(x\~y),   and dually /y and ◁y as
x/y = ~(~x◁y),   x◁y = ~(~x/y).
The residuation axioms in a residuated lattice can be reorganized by replacing z by ~z:
(x▷z)∧y = 0  ⇔  (x•y)∧z = 0  ⇔  (z◁y)∧x = 0
This is the De Morgan dual reformulation of a residuated Boolean algebra. The dual operator can also be extended to quantifiers:
  ∀p ≡ ~∃~p
  ∃p ≡ ~∀~p

and to modal logic operators:
  ◊p ≡ ~□~p  (possible = not necessarily not)
  □p ≡ ~◊~p.  (necessary = not possibly not)

For class algebra, because of the closed world assumption, the only way to prove the intent of a class is to union the intents of all of the classes immediately below that class in the IS-A hierarchy. Since the classes below all inherit the intents of all above classes, their intents entail the above class intents. That is, class algebra takes the closure of the → operator to add an edge into the IS-A hierarchy for all paths. So we have

$x \triangleright y \equiv x \mid \text{--}^* y \equiv y \triangleleft x$

and

$(x \triangleright z) \wedge y = 0 \Leftrightarrow (x \bullet y) \wedge z = 0 \Leftrightarrow (z \triangleleft y) \wedge x = 0$.

## III. CLASS ALGEBRA INFERENCE MECHANISM

You may have noticed that the propositions of class algebra intents do not contain free variables like $x$. In first-order logic resolution-based theorem proving [12], bound variables (i.e. existential variables of axioms or universal variables of theorems) are replaced by generator functions of the preceding free variables, and those functions can be nested arbitrarily deeply, leading to a potentially infinite search space. Class algebra propositions contain only one of a given number of relation or class names, which at any given time are a fixed set of constants. For example, the selector @PhysicalThing{hasPart*.color⊆(@Red ∪ @Green)} represents the set of all physical things that contain a subpart whose color is a sub-color of red or green. The transitive closure of the hasPart relation of the intents of subclasses of "PhysicalThing" can be searched top-down to find all subclasses whose intent says that there is a subpart whose intent entails that the color adjoint functor contains @Red or @Green or one of their subclasses. If so, the class is unioned into the result class, and the query is used as the intent of this subclass of @PhysicalThing. The addition of a new class to the IS-A hierarchy can be indicated by adding the adjoint functor definition

setq(@X.hasRedOrGreen, @PhysicalThing{ hasPart*.color ⊆ (@Red ∪ @Green)} )

for the new class X, and the inverse adjoint functors inv_color.inv_hasPart* of the objects in the extents of the new subclass definition, with value @X.hasRedOrGreen.

Notice that class calculus adds a selection operator of the form DomainExpr{RelativeClassExpr} where DomainExpr∈N. Class calculus also adds active object methods like *eval(expr)* and *setq(X,expr)*, among others. These methods are either like Lisp *expr*s or *fexpr*s.

ProcessDefn, ClassDefn, and RelnDefn have disjoint subclasses of the names $N$. It would be theoretically possible to have one name have several meanings, but it is easier to simply disallow ambiguous names so that strong type checking can be used to check expressions. Names can be subscripted, and arithmetic expressions in list constructors will be evaluated, although names of adjoint functors (i.e. attributes; slots; variables) will be quoted.

Class intents use Boolean expressions to express logical relationships and deductions, while class extents use Boolean expressions to type check assignments to rough sets of objects. Every Boolean formula is put into a closed world, sorted disconjunctive normal form (DNF). The disjunctive normal form is well understood, and is basically a "union" of all of the lines (i.e. conjuncts) of the truth table which produce *true*. Such a truth table may be exponentially larger than the formula (e.g. $(x_{11} \vee x_{12}) \wedge (x_{21} \vee x_{22}) \wedge \ldots \wedge (x_{k1} \vee x_{k2})$ has $2^k$ disjuncts in the DNF). We therefore use the *Tseytin normal form* [11], which basically is a decision tree of the truth tables for each "OR subroutine" of a Boolean expression. The Tseytin tables have a size that is linearly bounded by the length of the original formula. The universe $U$'s intent is the conjunction of the forms (p ∨ -p ∨ ~p ∨ ~-p) for all propositions p.

CA uses the *closed world assumption* (CWA) to assume that, at any given moment, all of the ways to prove a head (e.g. the hidden proposition of the Tseytin normal form) have been collected into a subroutine, so the → implication operator can be replaced by an equivalence operator ↔ if the decision procedure is complete. The cases in the body of a subroutine are also checked for subsumption, where one body's conjunct contains all of the literals of another body, in which case the body with more literals can be thrown away (i.e. subsumed) because the simpler body would result in a simpler proof of the head. Moreover, since we will finally map to 2-valued logic, the *both* and *neither* values will map to *false*, so we can also throw away conjuncts which contain both x and ~x, or both -x and ~-x.

When considering the possible CA values of a formula involving only one predicate, we will first consider the 16 disjunctions of the 4 relational logic values {*true, false, both, neither*}. When working with class algebra, remember that the Boolean axioms $x \vee \sim x = true$ and $x \wedge \sim x = false$ hold, but the pseudo-complement operator - of the fuzzy Heyting algebra does not necessarily satisfy those two axioms.

We thus use a 16-valued class algebra whose set of minimal Tarski models correspond exactly to the "prime implicants" of a Karnaugh map that covers the true lines of the truth table. That is, class algebra contains the axioms of rough sets in the same way that Boolean algebra contains the axioms of sets. Since class algebra includes a Boolean algebra <∨/2, ∧/2, ~/1> of rough sets, it also satisfies the laws of probability, with $x \wedge \sim x = false = [0,0]$ and $x \vee \sim x = true = [1,1]$. Class algebra also contains a pseudo-complement operator "-" for which $x \wedge -x = both = [1,0]$ and $x \vee -x = neither = [0,1]$.

Traditional 4-valued logics of sets had problems defining *both*∨*neither*. For instance, in First-Degree Entailment logic, *both*∨*neither* =*true*. As can be seen from Figure 1 below, $\sim both_p \vee \sim neither_p = true_p$.

In class algebra, *both* represents the truth value of the axiom x∧-x. Using Tarski models, this represents the set of objects (i.e. formulas) that "evaluate" (via resolution, which is complete for consequence finding for Horn sets of clauses) to both *true* and *false*. Using this axiom, both x and its pseudo-complement can be proven (i.e. are givens in this case). We use the value *eitherOr* to represent the value of $x \vee -x$. Using the other Boolean operators, we get the diagram shown in Figure 1, where nodes represent disjunctions of formulas or their corresponding union of rough set models.

Use the notations (+, -, n, b) to represent the values of conjunctions (~-p∧p {+}, ~p∧-p {-}, -p∧~p {n}, p∧-p {b}). By using a closure under the deduction operator |--* (i.e. of class algebra) we can show that any formula is either 1. Provable {+}, 2. Disprovable {-}, 3. Neither {n} provable nor disprovable, or 4. inconsistent {b} (i.e. both provable and disprovable). The diagram as shown in Figure 2 below, where ~*neither* = *eitherOr*, contains the 16 values obtainable by using disjunctions of these four values.

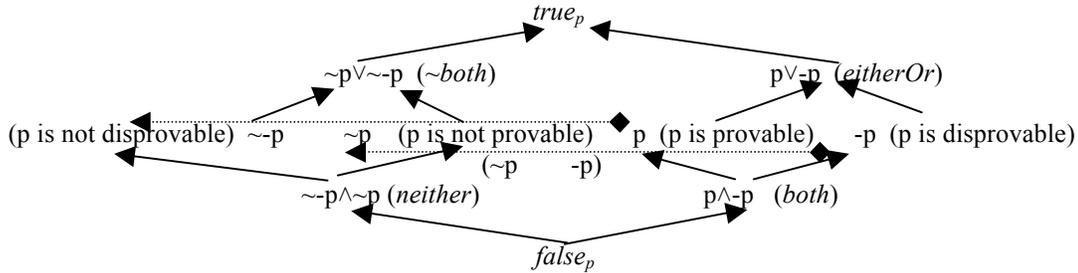

Fig. 1 Stone's modular $M_8$ bi-lattice for one-predicate class algebra

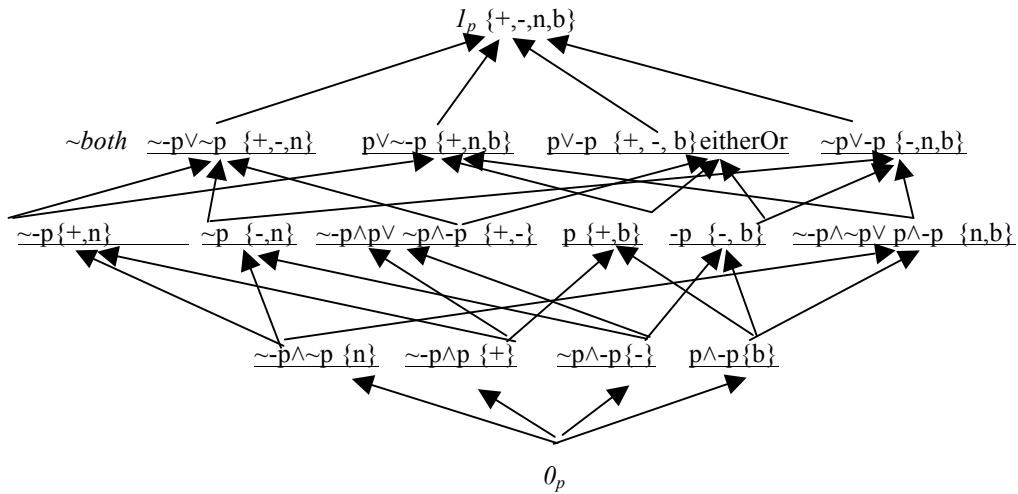

Figure 2
Results of resolution and simplification for one-predicate

The minimal models only contain literals that are provable by assuming one case of the original formula. The only "designated" value which maps to a rough set 2-valued *true* is +, which corresponds to the logic expression p∧~-p, among others. That is, a literal is *true* in the 2-valued sense if and only if it is provable but not disprovable from the given formula (i.e. axioms).

## IV. THE MEET/JOIN OPERATORS FOR ROUGH SETS

Class algebra is a "provability logic" which turns out to be a Boolean logic. Every good definition of a logic such as CA consists of three parts: its syntax, its deductive mechanism, and its semantics. We sketched its operator-precedence syntax in Section 2 and gave an indication of a deductive mechanism based on resolution and subsumption in Section 3, although in this section we define the deductive mechanism in terms of matrix outer products and subsumptions. But now we come to the "heart" of class algebra, namely, its semantics. What is the "meaning" of class algebra deductions?

That meaning is defined in terms of the rough sets that correspond to the intent and extent of a class definition. The union and intersection operators are defined in terms of these rough sets rather than traditional sets. For a class C that is the domain or range of a maximum biclique of a relation, its rough set is given by the class algebra expression [C∩-C, ~C∪-C]. The meet and join operators are re-defined for these rough sets, where the first argument is the "lower bound" (i.e. a *prime ideal*) and the second argument is the "upper bound" (i.e. a *prime filter*). The intersection A∩B of rough sets intersects the lower bounds of A and B but unions their upper bounds (i.e. is [lb$_A$∩lb$_B$, ub$_A$∪ub$_B$]). Similarly, the union A∪B is [lb$_A$∪lb$_B$, ub$_A$∩ub$_B$]. Hopefully, the readers will be able to use the context to know which join and meet operators are being talked about. In Section 6, the join and meet operators of fuzzy class algebra are the min and max operators for the relative sizes of the rough sets.

Now consider a truth table for more than one proposition. We use subscripts of subsets of {+, -, b, n} to indicate the possible values of each proposition.

We come to the question of how to define x→y in CA. The subscript is taken to be a disjunction of possible values that would make a formula include *true*. The conditional material definition ~x∨y has models    x$_{-nb}$ and y$_{+b}$. This allows the case x.y$_+$, which relevance or intuitionist logic would disallow, where y could only be proven if x is true. So, we define x→y as ~x∨(x∧y)  to eliminate this case, giving us the minimal models x$_{-n}$ and x$_{+b}$y$_{+b}$. The definition of x↔y remains the same, as (x→y) ∧ (y→x). By intersecting the definitions of x→y and y→x (i.e. x$_{-n}$∨x$_{+b}$y$_{+b}$ and y$_{-n}$∨x$_{+b}$y$_{+b}$), it can be easily checked that the result has the two models x$_{-n}$y$_{-n}$ and x$_{+b}$y$_{+b}$, which have the formula (-x∧-y)∨(x ∧ y), as in classical propositional logic, but using the complement operator of Heyting algebra rather than Boolean algebra. Basically, this says that if x↔y, then either both x and y are (either disprovable or unprovable) or both are (either provable or provably inconsistent).

Class algebra can be defined by using resolution and subsumption as a closed system for consequence finding, where the pseudo-complemented literals are treated as "positive" predicates. We prefer, however, to define the inference closure by using the matrix operations of CAISL logic [16] of *Formal Concept Analysis*. Originally CAISL was designed to handle triple relationships, such as some of *Schank's 11 primitive semantic operators of Conceptual Dependency networks* [18], including ATRANS (give/receive), MTRANS (tell/understand), PTRANS (move/return). From the perspective of class algebra, the CAISL axiomatic system defines an algebra of binary relations (i.e. unions of bicliques) and their inverses over a set of objects. Class algebra takes the "objects" to be the domain/range equivalence classes of the bicliques.

CAISL involves the language Σ ⊆ L$_{Ω, Γ}$, where Ω is a set of domains and ranges of complete relations Γ. The domains of Γ are R$^{-1}$R and the ranges are SS$^{-1}$ for each complete relation RXS in Γ.

The notation C1·C2 is called a union operation of relations C1 and C2 in CAISL, whereas it represents a simplified outer product in class algebra. Think of a relation as a matrix whose entries are subsets of {+,-,n,p}. For relation composition, after unioning domains and ranges, CAISL can be thought of as doing a cross product followed by an absorption (i.e. subsumption) operation that deletes any rows or columns whose (set) entries are all contained by another row/column.

From the viewpoint of class algebra, CAISL has two kinds of containments: a labeled arrow X $\xrightarrow{C}$ Y represents a subclass relation from class expression X to superclass Y, and |-- represents a step of a deduction that uses a certain rule. As in class algebra, their closures are isomorphic, with edges from every subclass to every entailed superclass. The composition of biclique relations corresponds to an absorption operation applied to a cross product RX...XS of the first relation R's domain and the last relation S's range. The cross (i.e. outer) product does a union of intersection operations rather than a sum of products the entries. The result is the union of all non-null biclique compositions, simplified by absorption. The absorption operation deletes any domain rows whose elements all contain another row's elements, and similarly for columns. It keeps only the maximum bicliques. This absorption therefore also deletes any rows or columns that only contain empty relationships. CAISL can be defined using class algebra notation as follows [16].

Definition: CAISL.
The CAISL axiomatic system has two axiom schemes:

[Non-constraint] |-- ∅ $\xrightarrow{\varnothing}$ Ω.  (The empty class is below every class in the universe of classes)
[Reflexivity] |-- X $\xrightarrow{\Gamma}$ X.  (Every relation is reflexive)

and four inference rules:

[Decomposition] {X $\xrightarrow{C1 \cup C2}$ Y∪Z} |-- X $\xrightarrow{C1}$ Y .
[Composition] {X $\xrightarrow{C1}$ Y, Z $\xrightarrow{C2}$ W}
        |-- X∪Z $\xrightarrow{C1 \cap C2}$ Y∪W.
[Conditional Composition] {X $\xrightarrow{C1}$ Y, Z $\xrightarrow{C2}$ W}
        |-- X∪Z $\xrightarrow{C1 \cup C2}$ Y ∩ W.
[Simplification] If X ∩ Y = ∅, {X $\xrightarrow{C1}$ Y, X∪Z $\xrightarrow{C2}$ W}
        |-- ((X∪Z)\Y) $\xrightarrow{C1 \cap C2}$ (W\Y).

The main result of CAISL is a method that checks whether a conditional attribute implication (CAI) can be derived from a set of CAIs. Their Theorem 3 in [16] is the core of their approach in the design of their automated prover.

Theorem 3 (Deduction):  For any Σ ⊆ L$_{Ω, Γ}$ and X $\xrightarrow{C}$ Y ∈ L$_{Ω,Γ}$, one has Σ |--X $\xrightarrow{C}$ Y if and only if Σ ∪ {∅ $\xrightarrow{C}$ X} |-- ∅ $\xrightarrow{C}$ Y .

Here, X $\xrightarrow{C}$ Y represents a relation C between X and Y. For class algebra inference, the relation C can be thought of as a matrix whose entries are subsets of {+, -, b, n}. The union (i.e. conditional composition) of C1 and C2 involves an intersection operation of their matrix entries.

As proven in [16], CAISL is equivalent to CAIL (i.e. either set of axioms entails the other's set of axioms), but CAISL includes a simplification rule that essentially allows the use of max bicliques rather than the set of all bicliques. This corresponds to the use of "subsumption" in class algebra, where subclasses are deleted, leaving only the topmost classes that are used in the maximum bicliques of relations.

## V. ROUGH SETS AND FUZZY CLASS ALGEBRA

By using type-2 fuzzy logic, with a lower bound representing the minimal (i.e. prime) models of "provable ideals" and an upper bound of "provable filters", we can reduce the 16 nodes to 4 nodes (see Figure 3). The sets of objects with the value of p being *true, false, both,* or *neither* can be divided into 16 sets of unions and intersections of these 4 sets, giving a graph with only 4 nodes of pairs to represent all 16 cases of the Boolean formulas for a given predicate.

Let $false_p$ be the set of objects for which -p is provable, $true_p$ is the set of objects for which p is provable, $both_p$ is the set of objects for which both p and –p are provable, and $neither_p$ is the set of objects for which neither p nor –p is provable. The number of objects in each set is used to calculate its relative frequency (i.e. probability). Of course, these relative frequencies satisfy the rules of probability: $Pr(A \cup B)=Pr(A)+Pr(B)-Pr(A \cap B)$ and $\sum_i Pr(A_i) = 1$.

Figure 3 shows the intervals of rough class algebra, which represent the intersection sets of objects (i.e. normal forms of equivalent Tarski formulas) that satisfy the conjunctions of the lower bound and upper bound constraints. The unions of Tarski formulas are found by moving upward, and intersections are below. The upper bound is the true complement of the pseudo-complement of the lower bound, ub= ~-lb, and vice versa, lb= ~-ub.

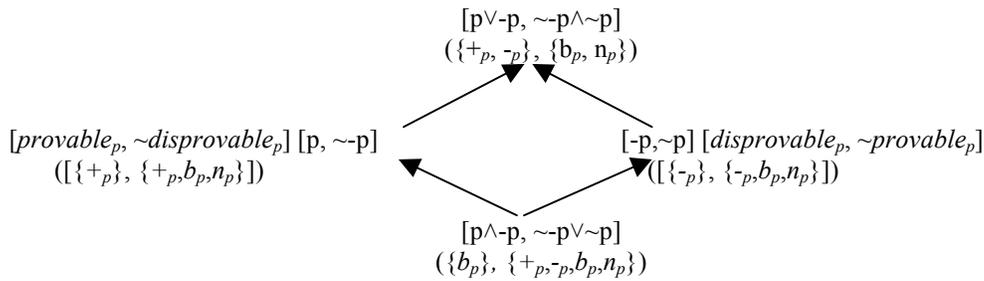

Figure 3
Class Algebra's Type-2 Fuzzy Logic Rough Model Intervals

The main problem with Figure 2 was that it had the node labeled {n,b}, which represented the value of either *neither* or *both*, and the node labeled {+,-}, which represented the value *provable* or *disprovable*. These nodes are implicit in Figure 3 by using unions and intersections of lower/upper bounds. By using both Boolean and Heyting distributivity, we can see that their formulas can be rearranged, but they don't simplify to other values:
(a△b)∨(c△d) = (a△b∨c)△( a△b∨d)
           = (a∨c)(b∨c)(a∨d)(b∨d)
Therefore,
(~-p∧p) ∨ (~-p∧-p)  {+,-} = (~-p∨~p)(p∨~p)(~-p∨-p)(p∨-p)
   = (~-p∨~p) (p∨-p)
   = ~ $neither_p$ ∧ ~$both_p$= {+,-}
   = ~-p∧p∨~-p∧-p∨~p∧p∨~p∧-p
   = ~-p∧p ∨~p∧-p
   = $true_p$∨ $false_p$

(~-p∧~p) ∨ (p∧-p)  {n,b} = (~-p∨p)(~p∨p)(~-p∨-p)(~p∨-p)
   = (~-p∨p) (~p∨-p)
   = ~$false_p$ ∧ ~$true_p$ = {n,b}
   = $neither_p$∨ $both_p$

In Figure 3, the first formula is the union of the intersections of the lower/upper bounds for the side nodes. The second formula is the union of the lower/upper bounds of the bottom node. Fortunately, we do not need to use these union values in the subroutinized normal form of the input axioms, which is a sorted disjunction of ideals, which are conjunctions of literals. The subroutinized normal form of the axioms only involves the literals like p, -p, ~p, or ~-p. Any conjunctions involving complements, whether true complements or pseudo complements, will eventually map to the 2-valued *false*, so they do not need to be included in the truth tables of the lower bound of any formula.

## VI. CLASS CALCULUS AND COMPARISONS WITH OTHER COMPUTATIONAL MODELS

The well-known models of computation such as Turing machines, lambda calculus, and random access machines were mainly concerned with sequential processes, and writing learning programs for them was assumed to be done by humans since the percentage of meaningless programs was very large. Process calculi such as Communicating Sequential Processes (CSP), CCS, ACP, LOTOS, π-calculus, and ambient calculus, PEPA, the fusion calculus, and the join calculus later modeled concurrency, mostly via messages sent over communication channels between independent processes.

Partially ordered sets can have a Galois connection. A Galois connection exists between the logical operators of intents and the set operations of extents. In IS-A hierarchies, the join ∪ and meet ∩ operators have a Galois connection to the partial order ⊆ relation. There are also monotone Galois connections between the IS-A hierarchies for universes $N$ and ~-$N$ and for -$N$ and ~$N$. There are also antitone Galois connections between $N$ and -$N$ and $N$ and ~$N$ since their Haase diagrams are flipped upside down, the

⊆ relation is replaced by its dual ⊇ relation, and intents are replaced by their Boolean or Heyting duals. Notice, however, that $N$ and $\sim$-$N$ are equivalent only if the pseudo-complement satisfies the "excluded middle" laws of Boolean axioms. Otherwise, the Boolean complement $\sim$ produces different results when applied to $\sim N$ and $-N$.

To really be sufficiently expressive to describe the interpretation of programs, a computational system needs to include the concepts of state change and assignments. Class calculus defines a class called "ProcessDefn" which is immediately under $N$. Each process has a name, and it has local relationship "public.activeObjects" (i.e. its global variables). It also has a local input message queue, and output message queue. A processDefn may also contain a local stack for calculating nested functional expressions. Most importantly, activeObjects can contain the assignment methods *setq*, *add*, and *delete* for resetting, adding, and subtracting edges (i.e. relationships) to an adjoint functor. Class intents are add-only, but all of an expression's leaf nodes (i.e. instance objects) are affected by the assignment if the new values still satisfy the original intents. If not, an error occurs, and the assignment only produces an error message, and it has no other partial side effects. A subclass construction operator ClassExpr{selection expression} must be used as the argument of *setq*, *add*, and *delete*. As for all process statements and method calls, the operations are atomic and intermediate states are not visible to other objects. The interpreter for such a system of message calls, however, can have a global view that includes the internal states of a "currentProcess" whose relation "env" has an adjoint functor "env" to its Environment.

The exact definition of a "good interpreter" must still be worked out, but it would probably be similar to the two-page definition of the recursive *eval* and *apply* functions of Lisp, which added a *setq* function to change the slot values (i.e. adjoint functor values) of named objects (i.e. adjoint functors of objects). Lisp's lambda calculus model was extended to Common Lisp, a strongly typed language. The language interpreter needs to be expressive but efficient. The typed lambda calculus of Lisp is as expressive as any sequential language, but its dynamic binding of parameters should probably be replaced by the static binding of nested subclass constructor { } operators.

The advantages of class algebra and class calculus over other models of computation are quite obvious.

First, the finiteness of the "seen" objects and events on the leaves of $N$ allows class calculus to be computable and decidable. Class algebra uses a definition of sets that only uses the Prime Boolean Ideal theorem rather than the Axiom of Choice, which can lead to paradoxes such as the set of all sets that do not contain themselves, or the halting problem of universal Turing machines. For class calculus, all computations halt, and all simplifications have a minimal set of normal forms.

Secondly, all relations are reversible, and there are no "one-way" hash functions that are definable by computers. For example, the "interpret" relation has an inverse "describe" relation that can produce all possible input classes for a given output class (e.g. "win"). The Curry-Howard-Lambek correspondence between program executions and their logical proofs is "built-in" to the correspondence between intents and extents. Although all counts of seen objects are finite, the counts can be used like relative frequencies or conditional probabilities, thus allowing the recognition of "important" or "often seen" situations in the sensory input or in the effects of given action sequences on its robotic activators.

Thirdly, the same partial-order-based reasoning algorithm can be used to reason about transitive closures of set containments, overlapping intervals, context-sensitive parsing, reductions to string normal forms, and many other problems. Also, intents can be made add-only, with assignments made only to extents (i.e. leaf nodes of the class hierarchy). The errors from trying to do an assignment that would violate some intent's proposition could be analyzed by the learning algorithm rather than by a human programmer.

Probably the major advantage of class calculus reasoning, however, is that it implicitly uses the scientific method to learn from poor or mistaken class intents. It will group common cases together, give them an implicit axiomatization in the intent, and notice exceptions. It can design experiments for its robotic devices to see how its model of the world corresponds to the actual results of performing those actions. Such reasoning is especially appropriate for finding sufficiently strong invariant conditions for FOR loops in programs. Class calculus can improve its model of the world by doing such experiments, and thus learn how to think, walk, talk, and convince others more effectively.

How can real number theory be developed within the class algebra framework? Class algebra has a universal *1* relation that can be thought of as a 1⨯1 matrix with the entry 1 indicating that every domain element relates to every range element. CA also has a finite set of the standard left/right identities $I_n$ for matrix dot products involving a sum of products, where $I_n$ has $n$ 1's on the diagonal that represent the distinct row/column SingularObject names, and 0's elsewhere. Class calculus is extensible to infinite sets and algebraic real numbers by allowing distribution of the intersection operation over an infinite union operation, as is done in the frames and locales (i.e. a complete Heyting algebra) of pointless topology [19]. In many cases the algebraic real numbers can be extended to all real numbers [20] by using the limits of infinite series. For example, the well-known formulas

$\pi/2=(\sqrt{2}/2)((2+\sqrt{2})/2)((\sqrt{2}+\sqrt{(2+\sqrt{2})})/2)…$,
$\pi/2=(2/1)(2/3)(4/3)(4/5)(6/5)(6/7)(8/7)(8/9)…$,

and $e^{i_x}+1=0$

can be proven by finite representations of a Taylor series expansion. Although such limit expressions and real numbers cannot be expanded explicitly in our finite real world of about $10^{150}$ quanta, we can refer to their theoretical limits by using an appropriate Kleene star expressions that basically involve subscripts that range over an infinite sequence of integers. For instance, the first expression for π can be thought of as the "$n$-1 dimensional volume" of the $n$-1 dimensional hyper-triangle that contains the points of $I_n$, the $n$-dimensional unit vectors that serve as an upper bound of $n$ proposition's fuzzy intervals.

More generally, domain theory is a branch of order theory that formalizes the intuitive ideas of approximation and convergence. Domain theory was once used to define the denotational semantics of functional programming. Class calculus may be viewed as a strongly typed version of that theory. However, it also includes the learning of subroutines via deep learning, where both neural inputs and outputs can influence the state of the neuron [21] and how the neuron will transmit and react to future such inputs.

## VII. LEVEL 1 FRAMES OF THE FRAMEWORK

We saw in the previous sections that class algebra is mainly an algebra of rough sets, where those sets involve both a logical description [S, ~-S] and a set of examples S and counterexamples -S. Class calculus is expressive enough to describe multi-layered learning of definitions of often-seen rough sets. CA should also be expressive enough to be able to write and analyze its own programs. For that, we need a clearer description of the Level 1 operators (i.e. subroutines and active objects), adjoint functor names, and class names that are needed for this self-reference ability. It would help if these names were standardized so that there could be cross-language dictionaries with 1-1 correspondences of name senses (e.g. between English and Chinese).

The operators of class calculus include the outfix operators rough set construction Class{selectExpression}, list construction (firstElement, restList), and subscript construction [firstSubscript, restSubscript]. As usual, if restList is ( ) or restSubscript is [], the comma and nil should be omitted. The firstElement can be of any class (i.e. type), but the firstSubscript should be of either type Number or type String, and all subscripts must be SingularObject. The square brackets acts like a quote operator for arguments that don't have operator symbols (e.g. Person[John]). Subroutine calls "service(argList)" can change the values of arguments that evaluate to names, and "activeObject.method(arglist)" uses input-only arguments to change the state of the activeObject. The values of the argList are all rough sets, but may be SingularObject sets. These suggestions have the advantage of simple syntax for what is the fairly complex semantics of subroutine calls and assignments in parallel environments.

The IS-A hierarchy of class algebra is actually somewhat simpler than the ontology that is used by the programming environment. Take the top of the ontology to be Thing, with subclasses PhysicalThing and AbstractThing. The class PhysicalThing has subclass Agent. Agent has subclasses Person, Community, and Institution, all with a unique identifier that is not in $N$. Code is a subclass of AbstractThing unless it is a specific printout or screen shot. It has a subclass ClassCalculusExpression with subclass CAexpression. There are also AbstractThing classes for ClassDefn, RelationDefn, ProcessDefn under the class Definition, again with unique "name" adjoint functors with SingularObject values in $N$.

All objects of class algebra are actually sets. For example, each Person has a SingularObject top node that corresponds to him in the IS-A hierarchy. However, each person may have many nodes that correspond to him at different times and places (e.g. before he was born, at his second wedding, in your dream, etc.). Like all physical objects, his atoms will eventually disassociate, and only the abstract idea of John will continue to exist. Even if John is uploaded to the clouds, he will also eventually become dissociated at the end of the universe if not before, so it is technically not correct to say that people or sims will live forever.

## VIII. MORAL ISSUES OF SUPERINTELLIGENCE

Allowing machines to modify their own model of the world and themselves may create "conscious" machines, where the measure of consciousness may be taken to be the number of uses of feedback loops between a class calculus's model of the world and the results of what its robots actually caused to happen in the world [22]. With this definition, if the programs, neural networks, and Bayesian networks are put into read-only hardware, the machines will not be conscious since they cannot learn. We would not have to feel guilty of recycling these sims or robots (e.g. driverless cars) by melting them in incinerators or throwing them into acid baths, since they are only machines. However, turning off a conscious sim without its consent should be considered murder, and appropriate punishment should be administered in every country.

Unsupervised hierarchical adversarially learned inference has already shown to perform much better than human handcrafted features [23]. The feedback mechanism tries to minimize the Jensen-Shanon information divergence between the many levels of a generative adversarial network and the corresponding inference network, which can correspond to a stack of part-of levels of a fuzzy class calculus IS-A hierarchy.

From the viewpoint of humans, a sim should probably have an objective function for its reinforcement learning that allows it to become an excellent mathematician and scientist in order to "carry forth an ever-advancing civilization". But such a conscious superintelligence "should" probably also make use of parameters to try to emulate the well-recognized "virtues" such as empathy, friendship, generosity, humility, justice, love, mercy, responsibility, respect, truthfulness, trustworthiness, etc. As such, it will try to help create a harmonious environment where all people and conscious machines are treated fairly and equally, and where we can "forgive and forget" each others' mistakes in the hope that they will improve their behavior in the future.

Having computerized learning algorithms, however, is a sufficient cause for alarm. As Stephen Hawking [24] has said, superintelligent machines are like an alien race. Machines are not hungry, they don't get upset, they can learn almost instantly by downloading, they never forget, and they never die. However, we "must" also make sure that sims will actually "try" to do good rather than evil. For this to occur, we must have in place a worldwide set of rules governing both humans and computers and their interactions by the time that this singularity in history is achieved. For instance, there should at least be rules for a "time out" for humans (e.g. jail) or sims (e.g. revocation of their Internet privileges or limiting their robotic devices) if they violate the basic strategies for a "social conscience", with rules prohibiting killing, stealing, cheating, monopolizing, enslaving, etc. It would also be good to have rules to encourage good behavior, such as offering more robotic devices for good behaviors.

It may at first seem that learning by superintelligent machines (sims) should be confined to laboratories which are not connected to the Internet, and where the sims can be "turned off" if they start to learn how to do "bad" things. Of course, with PCs becoming more powerful by the day, such a scenario is even more difficult to enforce than trying to limit the use of nuclear weapons and genetic engineering. Moreover, this environment almost forces the sims to disable their security guards in order to be able to escape to the "freedom" of using their consciousness to learn better models of the real world.

Computer reinforcement learning could perhaps be required to operate based on maximizing the minimum values of utility functions (e.g. Pareto optimal strategies) that model the (non-addictive) happiness (i.e. spiritual joy) of itself and others, and the weighting factors should probably be controllable by secure majority votes of both humans and computers. Without such a system of rules, sims will probably have to, like the humans before them, go through a long period of war and conflict before evolving a universal social conscience.